\documentclass[runningheads]{llncs}
\usepackage{graphicx}
\usepackage[utf8]{inputenc}
\usepackage{xcolor}
\usepackage{xspace}
\makeatletter
\DeclareRobustCommand\onedot{\futurelet\@let@token\@onedot}
\def\@onedot{\ifx\@let@token.\else.\null\fi\xspace}

\def\eg{\emph{e.g}\onedot} 
\def\ie{\emph{i.e}\onedot}

\def\etal{\emph{et al}\onedot }
\makeatother

\begin{document}
\title{Multimodal Semantic Scene Graphs for Holistic Modeling of Surgical Procedures}

\titlerunning{MSSG for Holistic Modeling of Surgical Procedures}
% If the paper title is too long for the running head, you can set
% an abbreviated paper title here

\author{Ege Özsoy\inst{1,}\thanks{Both authors share first authorship.}, Evin Pınar Örnek\inst{1,\star}, Ulrich Eck\inst{1}, \\
Federico Tombari\inst{1,2}, Nassir Navab\inst{1,3}}

\institute{
Computer Aided Medical Procedures, Technische Universit{\"a}t M{\"u}nchen, Germany
\and
Google
\and
Computer Aided Medical Procedures, Johns Hopkins University, Baltimore, USA 
}

% \authorrunning{F. Author et al.}
% First names are abbreviated in the running head.
% If there are more than two authors, 'et al.' is used.

%
\maketitle  % typeset the header of the contribution
\begin{abstract}

%% SHORT ONE:

From a computer science viewpoint, a surgical domain model needs to be a conceptual one incorporating both behavior and data. It should therefore model actors, devices, tools, their complex interactions and data flow. To capture and model these, we take advantage of the latest computer vision methodologies for generating 3D scene graphs from camera views. We then introduce the Multimodal Semantic Scene Graph (MSSG) which aims at providing a unified symbolic, spatiotemporal and semantic representation of surgical procedures. This methodology aims at modeling the relationship between different components in surgical domain including medical staff, imaging systems, and surgical devices, opening the path towards holistic understanding and modeling of surgical procedures. We then use MSSG to introduce a dynamically generated graphical user interface tool for surgical procedure analysis which could be used for many applications including process optimization, OR design and automatic report generation. We finally demonstrate that the proposed MSSGs could also be used for synchronizing different complex surgical procedures. While the system still needs to be integrated into real operating rooms before getting validated, this conference paper aims mainly at providing the community with the basic principles of this novel concept through a first prototypal partial realization based on MVOR dataset. 

\keywords{Surgical Procedure  \and Semantic Scene Graph \and Operating Room}
\end{abstract}

\section{Introduction}

% motivation 

\begin{figure}[hbt!]
\centering
\includegraphics[width=1.0\textwidth]{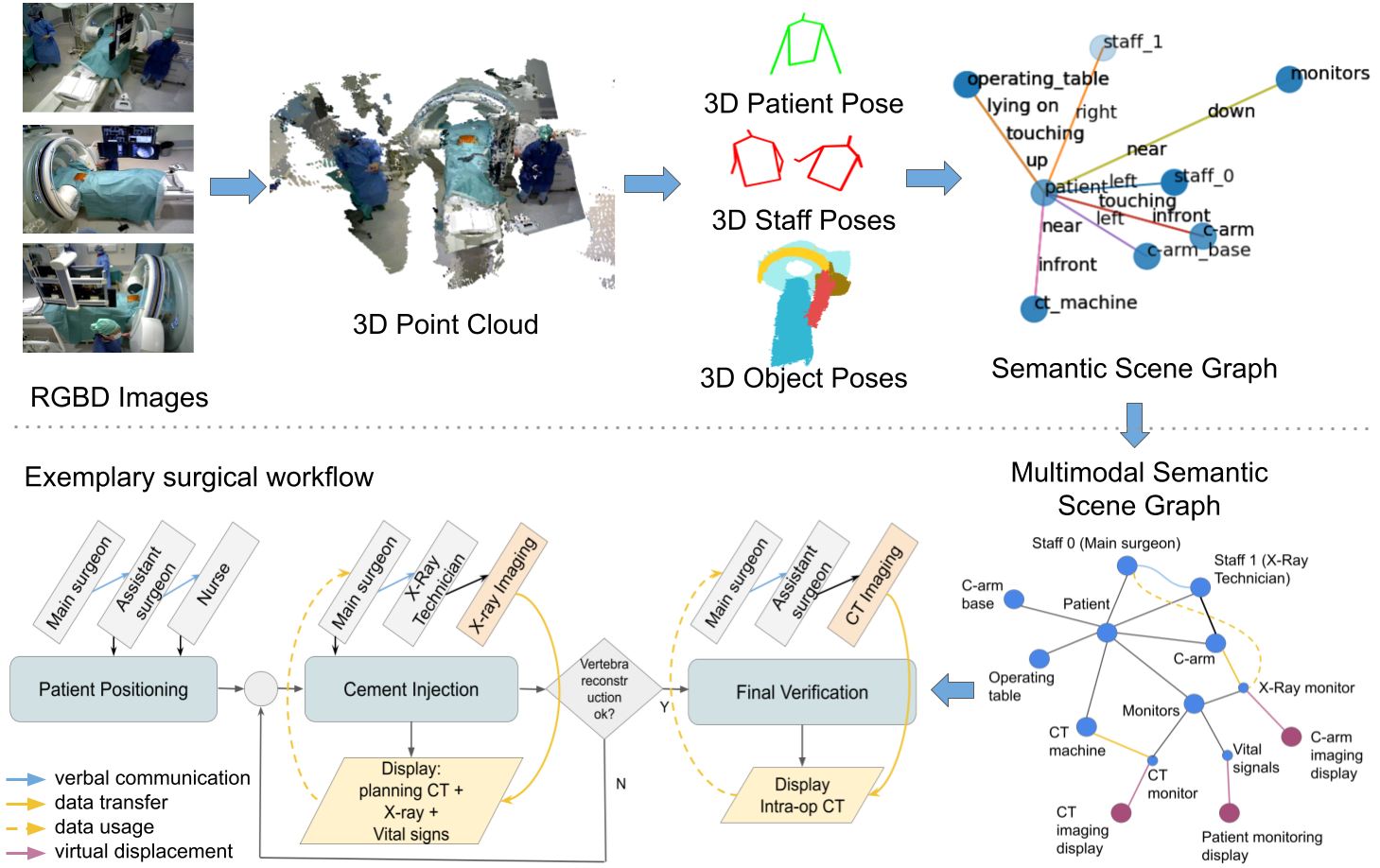}
\caption{Illustration of MSSG generation from multi-view RGB-D images. Top pipeline demonstrates the generation of SSG from MVOR dataset. On the bottom, the proposed MSSG is conceptually figurated and an exemplary case of vertebroplasty workflow modeling  derived from MSSG is presented.} \label{overall}
\end{figure}

Modeling surgical procedures with all their different components is crucial to develop intelligent systems that understand surgical operations, enable computer interventions, and provide situational aware assistance \cite{maier-hein_sds_2017}. Yet, as surgical procedures take place in highly complex, dynamic, cluttered and occluded environments, with the broad heterogenity of procedures, patients, medical teams, imaging and life support devices, surgical tools, operating table and monitors, they become a challenging input for visual perception and understanding \cite{kennedymetz2020,mohareri20203d}. Even though state-of-the-art computer vision techniques have proved to be highly accurate for task-specifically designed datasets, it is non-trivial to translate these methods to the highly complex real OR environments. Lalys and Jannin \cite{lalys2014} provide a review of surgical workflow analysis and establish the definition of Surgical Process Models (SPM). Furthermore, there is a great body of work on video and signal based surgical phase recognition \cite{garrow2020,nwoye2019weakly,czempiel2020,bodenstedt2020}, but the community has not yet focused on holistic approaches enabling the required dynamic mappings between the surgical process model, the human roles and the information processing and visualization flow of each of the devices within the operating room. This is however needed for digital systems to know who needs which information or tool from whom or which system at each moment of the surgery.

In this work, we introduce a concept and methodology for modelling and analyzing surgical procedures via multimodal 3D semantic scene graphs (MSSG). Semantic scene graphs (SSG) are introduced in the literature as a symbolic representation to connect visual attributes to semantic (\ie language-based) factors \cite{johnson2015}. They provide an abstraction layer in between those, and are specifically suitable for modeling the interactions between objects in complex environments. Typically, nodes of the graph represent the objects, and edges the corresponding relations between the objects. In the recent literature, SSGs are used for image retrieval \cite{johnson2015}, image generation and editing \cite{johnson2018,dhamo2020}, video understanding \cite{ji2020actiongenome}, 3D scan modeling \cite{wald20203dssg,armeni20193d}, and 3D dynamic simulation modeling \cite{rosinol2020dynamicsg}. 

SSGs proved to be convenient for modeling environments, but have not yet been studied in the context of highly complex environments such as OR rooms. Here we take SSGs one step further. We connect the various OR domain components and their multimodal data in a unified symbolic, spatiotemporal and semantic way through Multimodal SSG (MSSG). This representation allows adding virtual objects and connections between objects, connecting real objects to virtual ones and creating semantic relationships in between. Furthermore, this graph can be dynamically updated in real-time whenever a change is detected. One can adjust different levels of complexities depending on the use-case of the graph and the requirements (\eg patient focus, device focus).  

In the following, we first describe how to build a dynamic 3D scene graph specifically for OR environments, through the combination of various computer vision techniques for human pose estimation and object detection. We demonstrate our concept by means of the publicly available multi-view operating room dataset MVOR \cite{mvor2020}. We provide a suite of graphical user interface tools for surgical domain analysis, based on our symbolic representation of MSSG: \textbf{(a)} Bird's Eye View surgical process visualizer, and \textbf{(b)} text-based  report generator. 

Furthermore, we define a MSSG distance function enabling the measurement of similarity between two MSSG instances. We then use a simple Dynamic Time Warping algorithm, based on the proposed MSSG distance function to synchronize two surgical procedures. This demonstrates that MSSGs could offer many ways of using multimodal heterogeneous data for synchronizing surgeries, while putting the attention on different parameters or aspects of the surgical domain. For example, the surgeries could be synchronized based on the MSSG graph centered on the patient, surgeon or imaging device (see Fig. \ref{overall}) and in each case by putting the attention on different parameters based on the targeted application.

\section{Materials and Methodology}

\subsection{Scene Graphs}

A scene graph $G$ can be defined as a set of tuples $G = (N, E)$ where $N = \{n_{1},...,n_{n}\}$ is a set of nodes and $E \subseteq N \times R \times N$ is a set of edges with relationships $R$ \cite{johnson2015}. Nodes typically have a class label $c$ among different object categories (\textit{eg.} medical objects) or human types (\textit{eg.} specialist, patient). Edges have some relations, or attributes, $R$, from a set of different labels defining the relationships between nodes. In 3D scene graphs, nodes represent 3D objects \cite{wald20203dssg}. In order to build 3D scene graphs including human interaction, one needs to detect the object instances and humans in a given environment to be assigned as nodes and estimate the relationships between them to select edge labels. 

\subsection{Dataset}

In the light of the requirements to build a scene graph focusing on surgical operation room, there is only limited data published \cite{mvor2020,sharghi2020} to build a 3D scene graph, whereas only \cite{mvor2020} is publicly available. Multi-View Operating Room (MVOR) dataset \cite{mvor2020} provides 732 multi-view frames by three RGB-D cameras recorded in an interventional room during vertebroplasty and lung biopsy procedures in a 4 day setting. In order to create our system pipeline, we first fused the RGBD images from three cameras to build a 3D point cloud representation of the scene, by using the provided intrinsic and extrinsic camera parameters. The MVOR dataset provides 2D and 3D human pose annotations along with bounding boxes, however no annotations for objects nor patients. For human pose estimation, we use the existing labels for training a neural network, whereas for object detection we employ the classical computer vision algorithms \cite{ransac1981,icp1987}.

\subsection{Human Pose Estimation}
\subsubsection{3D Pose Prediction} Unlike other datasets where 3D scene graphs were applied \cite{wald20203dssg}, humans are central to our dataset to model the complexity of OR (\eg medical team and patients). Therefore, a robust detection and pose estimation of humans is very important. For detecting the 3D human poses, we selected VoxelPose \cite{voxelpose}, a state-of-the-art 3D pose estimation method. As this method requires 2D pose estimations as input, we employed HigherHRNet \cite{cheng2020bottom}, a state-of-the-art 2D pose estimation method. Both models are trained with the first three day data from MVOR, and validated with the last day.

\subsubsection{Patient Prediction}
The MVOR dataset does not provide patient pose labels, so it is not possible to involve patient prediction in human pose training. Instead, they provide bounding box labels which we used to train Faster-RCNN \cite{fasterrcnn2015} for patient detection from the images. On test time, to locate the patient in the point cloud, we apply this model to three images from three cameras. We crop out the parts that contain the patient, and project it to 3D. We then calculate the median of this point cloud for $(x,y,z)$, use this as the center of the patient and place a virtual patient pose model while always assuming the same position within the bounding box, head first, lying down orientation for the patient. 

\subsection{Object Recognition and Pose Estimation}

The MVOR dataset does not provide labels for objects either. Moreover, the dataset domain is significantly different than most indoor datasets that are commonly used in deep learning literature \cite{dai2017scannet} in terms of environment and presence of unknown objects. For these reasons, rather than depending on deep learning based detection and segmentation, we formulate our pose estimation task as \textit{detecting a 3D object scan from a 3D point cloud}, and focus on the objects that are detectable on MVOR data, such as monitoring devices and surgical machines. With the presence of a data with finer details, it would possible to detect smaller instruments along with different detection methods.

To achieve this, we manually extract the 3D point clouds of different objects visible in the environment. Once the object scans are created, the object pose estimation pipeline is employed. We iterate over the list of possible objects and apply a global finding and local fine tuning step: for every scan S of one object, we first calculate the normals, create downsampled version of it and calculate feature descriptors (FPFH, see \cite{fpfh2009}). Then we use Random Sample Consensus (RANSAC, see \cite{ransac1981}) to find the global location of this object. We assume the object is not in the scene if the RANSAC fitness is below a certain threshold which was determined manually specifically for MVOR. Then we apply colored Iterated Closest Point Algorithm (ICP, see \cite{icp1987}) to get a better alignment for the object. Once this has been done for every scan of an object, we use the pose with the highest RANSAC fitness value. 

%Object is added to the graph both geometric and semantic properties, and it's connectivity -> comes with the knowledge, ref fig \ref{overall}

% limitations, assume only one object, not very robust to changes on the object, works better on bigger objects

\subsection{Multimodal Semantic Scene Graph (MSSG)}

We propose the concept of MSSG, which extends the definition of SSG with virtual multimodal components to allow modelling the complexity of the OR environment. By \textit{virtual components}, we refer to the electronic signals, different outputs of the medical devices (\eg X-Ray, computational tomography, electrocardiogram, blood levels for anesthesia monitoring etc...), and display objects (\eg timer, BEV visualization display).

\subsubsection{MSSG nodes} The real OR components (humans and objects) are the first building blocks of MSSG, represented by graph nodes. Reconstruction and recognition of these components are described in previous chapter. The second type of nodes are the virtual components which can be added to MSSG to provide additional synchronized multimodal semantic information. 

Note that, objects present in the operating rooms could be defined together with their attributes and relationships. For example, as seen in Fig \ref{overall}, a C-arm would define an object that comes with a connection to X-ray technician. Once the C-arm enters in the room, graph should be updated with the C-arm node. C-arm object can have different states: on/off/providing acquisition/display transmission. An X-ray technician might turn on the C-arm, start acquisition, and the C-arm device can transfer the data to medical surgeon. Each of these steps can be represented via MSSGs.

\subsubsection{MSSG edges} There are different types of relationships among nodes: (a) spatial, (b) semantic, (c) virtual. The spatial relations are those that only describe the relative positional difference between two nodes, such as left, right, up, down, near or touching each other etc. The semantic relations carry non-spatial meaning, such as "getting scanned", "lying on", or "operating the patient on the torso". These are deducted using semantic information (\eg electric cautery on/off, coagulator on/off) and the spatial relationship between the nodes (\eg patient's relative position to the bed). Finally, virtual edges are those that describe the relations between virtual and real components, such as "CT scanning output (displayed on monitor 1)", "ECG signal input (displayed on monitor 2)".

\label{mssgdistance}
\subsubsection{MSSG distance function}
We define a distance function to be able to measure the differences between two given MSSG. Being a non-trivial task due to having different types of factors (spatial, semantic, virtual), we propose a simple function that counts all these factors, and can be modified according to user's objective. To calculate the distance between two scene graphs, we first retrieve elementary information about them separately, such as if they include any medical staff or a patient, how many medical staff is there, what objects are present in the scene, and object's relationship to the patient (\eg CT devices). These metrics are then compared between the two scene graphs and a distance is computed by weighting every difference according to a chosen set of factors, which can be listed as: \textit{presence of patient, medical staff, monitors, C-arm, patient monitoring, CT, endoscopic visualization, number of medical staff, presence of surgery on torso/head, position of patient and surgery acquisition time}.

\section{Surgical Procedure Monitoring through MSSG}

\subsection{Bird's Eye View Spatio-temporal and Semantic OR Visualizer}

A surgical procedure requires often many different devices and can quickly evolve towards a cluttered and complex environment, where one requires videos from different viewpoints synchronized with other signals to obtain a complete and informative overview during the surgery. For this reason, we propose a surgical operation visualizer with the help of MSSG which connects all components and provides a Bird's Eye View reflection with a simplified illustration. This visualization is created by projecting the scene graph nodes from top view. Virtual objects, such as patient transporter bed (not visible in any camera view), anesthetic data, X-Ray images, CT and its output, are also visualized.

\begin{figure}[hbt!]
\centering
\includegraphics[width=0.6\textwidth]{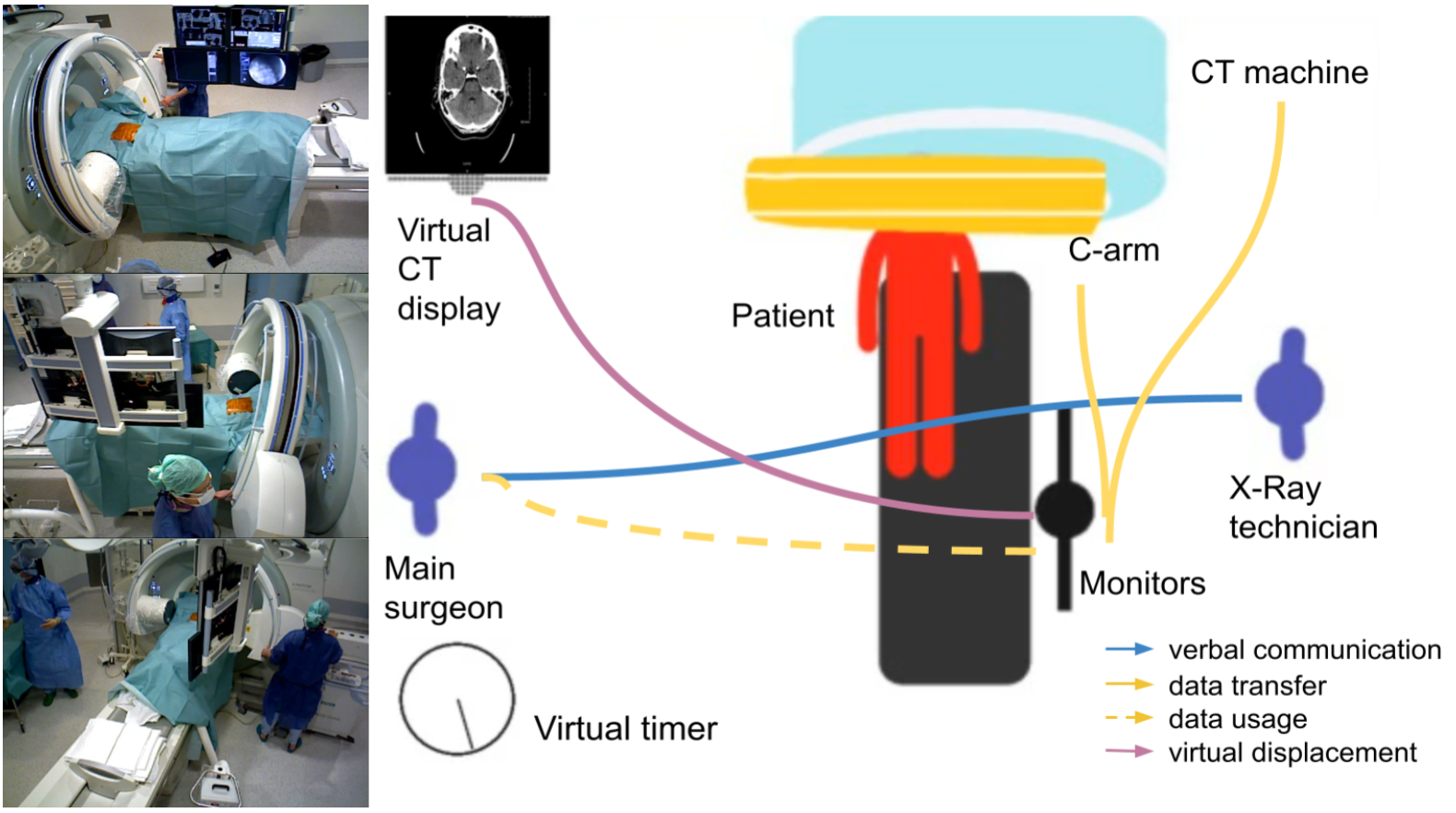}
\caption{Automatic Bird's Eye View  visualization of a surgical procedure through MSSG retrieved from 3 camera views. Visualization on the bottom left shows the C-arm, CT machine, patient, medical team, CT monitor and virtual CT display, virtual timer, along with conceptually proposed virtual edges.} \label{bev}
\end{figure}

%\subsection{Interactive 3D MSSG Visualizer}

%We provide an interactive MSSG visualizer, where user can select different nodes as a center of the graph, and visualize the connections and edge relations. For example, as most operations are patient-centric, one might select patient as a center and observe how other components are connected to it. 

\subsection{Event Change Reporter}

Our MSSG distance function allows comparing differences between two given scene graphs. As the distance is a function of semantic parameters, we provide an event change reporter which textually outputs the changes between two scenes. This can result in smart reporting and semantic modelling which could later be used for anomaly detection or augmented documentation.

\begin{figure}[ht!]
\centering
\includegraphics[width=0.6\textwidth]{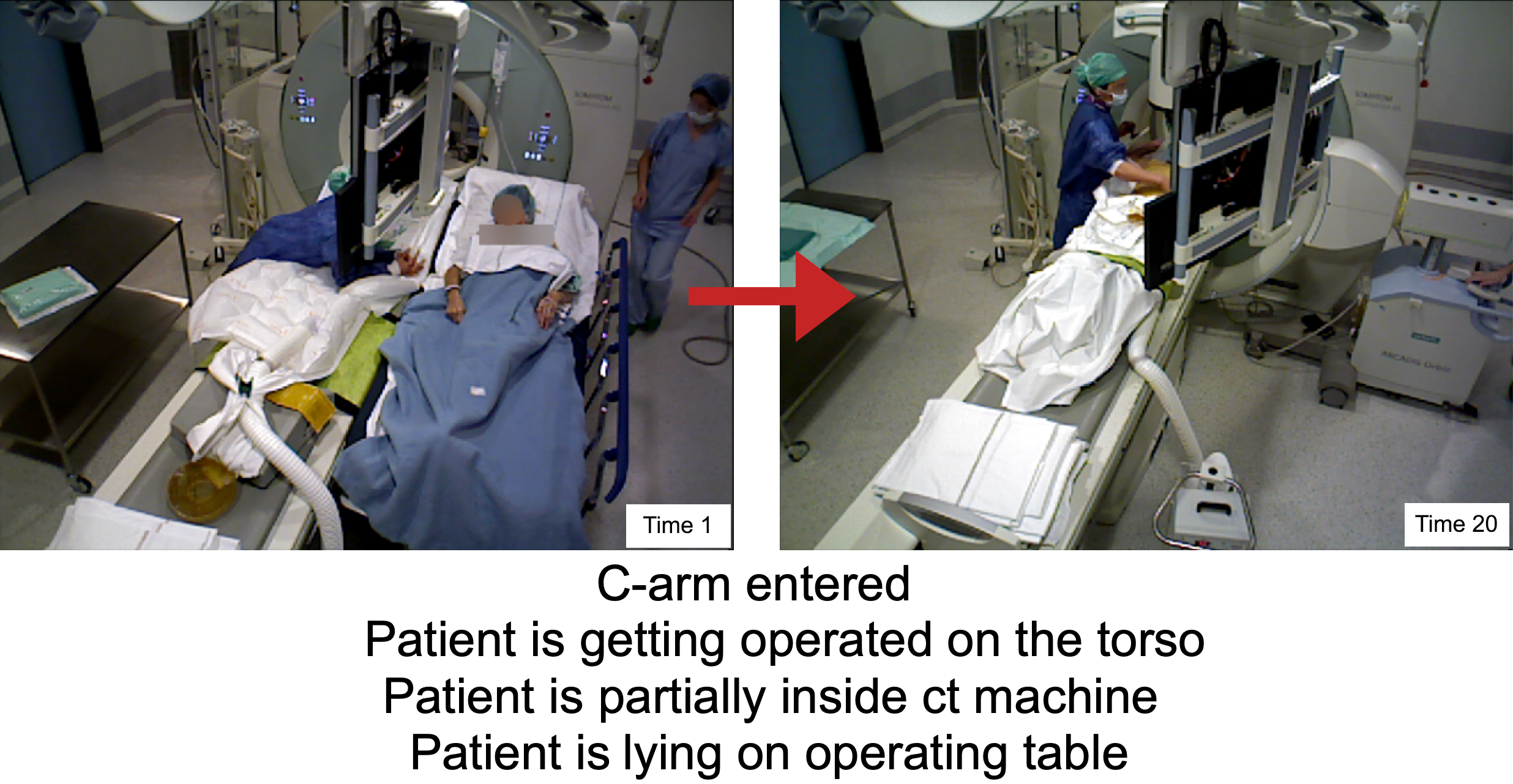}
\caption{Event change reporter outputs between two input scenes at time $t=1$ and time $t=20$, calculated from MSSG.} \label{reporterfig}
\end{figure}

\section{Surgical MSSG Synchronization}

\subsubsection{Problem Definition} We demonstrate one possible use-case of MSSG through the task of surgical operation synchronization, which aims to detect the corresponding events and landmarks between two given similar surgical operations \cite{FORESTIER2012255,ahmadi2006}. In our case, we can map the relevant phases of surgical operations simply through MSSGs. Our symbolic representation functions as a semantic abstraction layer allowing intelligent matching and synchronization, which would not be possible if only visual inputs were directly used.

\subsubsection{Dynamic Time Warping} One common and powerful method to measure the similarity between two temporal series is Dynamic Time Warping algorithm introduced by Bellman \etal \cite{bellman1959}. Given two series $a = \{a_1, ..., a_{m}\}$ and  $b = \{b_1, ..., b_{n}\}$, DTW aims to build a $MxN$ distance matrix $D$ which provides the distance for any path $P$ along the given sets \cite{bagnall2017}. The optimal path $P^{*}$ is the path with minimum distance, 

\begin{equation}
P^{*} = \min_{p\in P}(D_{p}(a, b))
\end{equation}

\noindent which can be calculated through dynamic programming.

In our setting, MSSG is the input for the algorithm, and the previously defined distance function is used to calculate the DTW distance matrix. Furthermore, as different attributes can bring more information to the synchronization, we applied the weighting scheme (see \ref{mssgdistance}) for distance function, where each attribute can have a different contribution on the overall distance (cost) function.

\subsubsection{Qualitative Results} 
\begin{figure}
\centering
\includegraphics[width=1.0\textwidth]{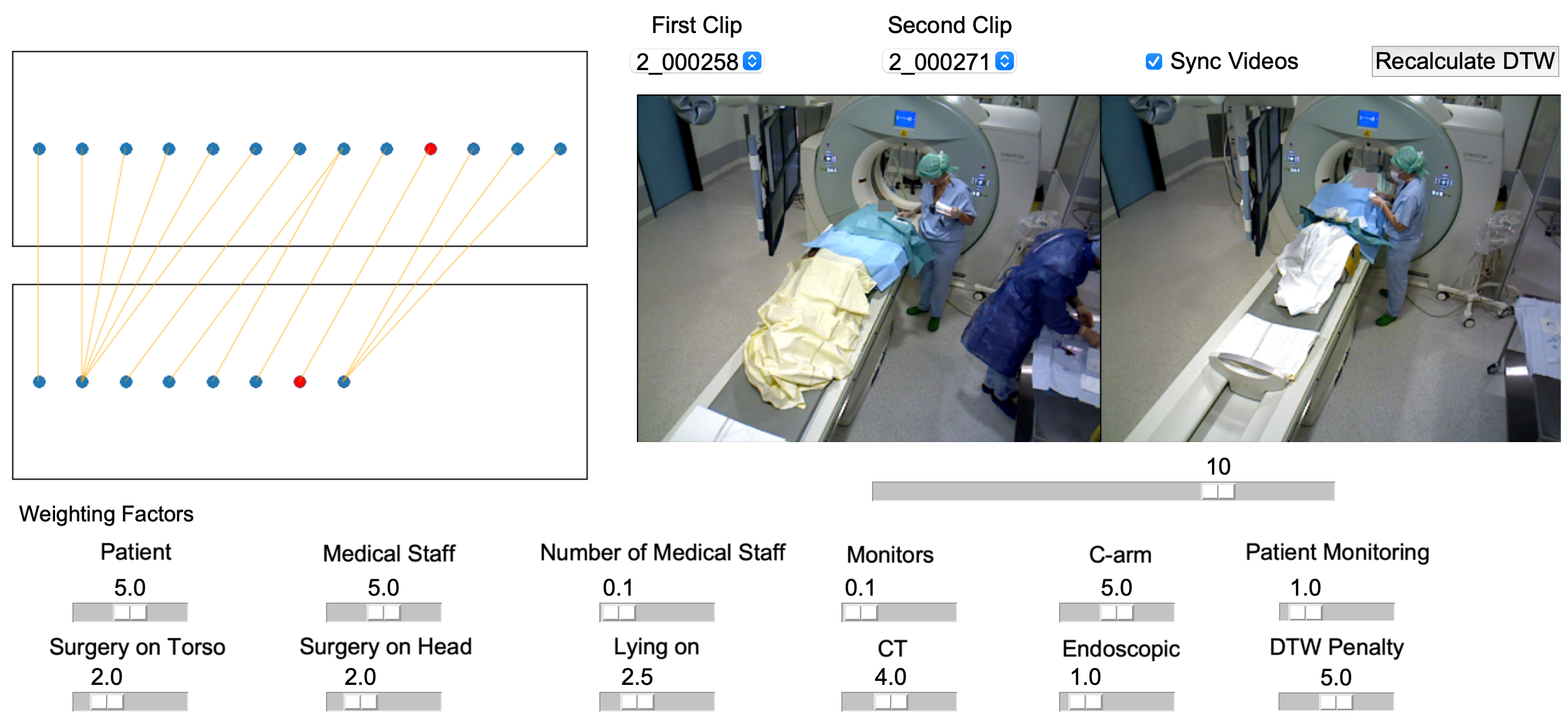}
\caption{Surgical MSSG synchronization tool based on DTW algorithm with MSSG distance calculated in a weighting scheme. Left boxes show the frames of video 1 and 2, where points show the matching between frames. On the right, the synchronized videos,  on the bottom, slider buttons allow selecting weighting factors. } \label{dtw}
\end{figure}

A GUI application that implements the scene graph DTW algorithm is provided, where a user can select the weighting mechanism and visualize the synchronization results between two given surgical videos(Fig. \ref{dtw}). 

\section{Discussion and Conclusion}

We proposed Multimodal Semantic Scene Graphs for modelling and analyzing surgical procedures in a holistic manner, which can connect various OR components in a semantically linked symbolic graph representation. The conceptual framework and a building pipeline for MSSG is described through different levels of visual recognition. A first set of surgical procedure analysis tools built on top of MSSG is provided, which allow visualizing an ongoing operation from Bird's Eye View, offering selection of different objects as the central node, and reporting the changes between two scenes. Moreover, surgical synchronization task is proposed through our MSSG distance based Dynamic Time Warping algorithm. 

The introduced system has many possible uses, such as medical staff role prediction, anomaly detection and smart reporting. Furthermore, more sophisticated OR tools and devices can be also fused into MSSG in real time and the semantic relationships can be enhanced through the prior knowledge and machine learning. Overall, this conference paper proposes a novel concept and creates new opportunities for learning from videos as well as synchronized signals acquired within operating rooms, opening the path towards holistic understanding and modeling of surgical procedures.

\bibliographystyle{ieeetr}
\bibliography{mybib}

\begin{thebibliography}{10}

\bibitem{maier-hein_sds_2017}
L.~Maier-Hein, S.~S. Vedula, S.~Speidel, N.~Navab, R.~Kikinis, A.~Park,
  M.~Eisenmann, H.~Feussner, G.~Forestier, S.~Giannarou, M.~Hashizume,
  D.~Katic, H.~Kenngott, M.~Kranzfelder, A.~Malpani, K.~März, T.~Neumuth,
  N.~Padoy, C.~Pugh, N.~Schoch, D.~Stoyanov, R.~Taylor, M.~Wagner, G.~D. Hager,
  and P.~Jannin, ``Surgical data science for next-generation interventions,''
  {\em Nature Biomedical Engineering}, vol.~1, p.~691–696, Sept. 2017.

\bibitem{kennedymetz2020}
L.~R. {Kennedy-Metz}, P.~{Mascagni}, A.~{Torralba}, R.~D. {Dias}, P.~{Perona},
  J.~A. {Shah}, N.~{Padoy}, and M.~A. {Zenati}, ``Computer vision in the
  operating room: Opportunities and caveats,'' {\em IEEE Transactions on
  Medical Robotics and Bionics}, pp.~1--1, 2020.

\bibitem{mohareri20203d}
Z.~Li, A.~Shaban, J.~Simard, D.~Rabindran, S.~P. DiMaio, and O.~Mohareri, ``A
  robotic 3d perception system for operating room environment awareness,'' {\em
  CoRR}, vol.~abs/2003.09487, 2020.

\bibitem{lalys2014}
F.~Lalys and P.~Jannin, ``Surgical process modelling: a review.,'' {\em
  International Journal of Computer Assisted Radiology and Surgery, Springer
  Verlag}, vol.~9, pp.~495--511, 2014.

\bibitem{garrow2020}
C.~R. Garrow, K.-F. Kowalewski, L.~Li, M.~Wagner, M.~W. Schmidt, S.~Engelhardt,
  D.~A. Hashimoto, H.~G. Kenngott, S.~Bodenstedt, S.~Speidel, B.~P.
  Müller-Stich, and F.~Nickel, ``Machine learning for surgical phase
  recognition: A systematic review,'' {\em Annals of surgery}, November 2020.

\bibitem{nwoye2019weakly}
C.~I. Nwoye, D.~Mutter, J.~Marescaux, and N.~Padoy, ``Weakly supervised
  convolutional lstm approach for tool tracking in laparoscopic videos,'' {\em
  International journal of computer assisted radiology and surgery}, vol.~14,
  no.~6, pp.~1059--1067, 2019.

\bibitem{czempiel2020}
T.~Czempiel, M.~Paschali, M.~Keicher, W.~Simson, H.~Feussner, S.~T. Kim, and
  N.~Navab, ``Tecno: Surgical phase recognition with multi-stage temporal
  convolutional networks,'' in {\em Medical Image Computing and Computer
  Assisted Intervention - {MICCAI} 2020 - 23nd International Conference,
  Shenzhen, China, October 4-8, 2020, Proceedings, Part {III}}, vol.~12263 of
  {\em Lecture Notes in Computer Science}, pp.~343--352, Springer, 2020.

\bibitem{bodenstedt2020}
S.~Bodenstedt, D.~Rivoir, A.~Jenke, M.~Wagner, M.~Breucha, B.~Müller-Stich,
  S.~T. Mees, J.~Weitz, and S.~Speidel, ``Active learning using deep bayesian
  networks for surgical workflow analysis,'' {\em International Journal of
  Computer Assisted Radiology and Surgery}, vol.~14, pp.~1079--1987, 2019.

\bibitem{johnson2015}
J.~Johnson, R.~Krishna, M.~Stark, L.-J. Li, D.~Shamma, M.~Bernstein, and
  L.~Fei-Fei, ``Image retrieval using scene graphs,'' in {\em Proceedings of
  the IEEE Conference on Computer Vision and Pattern Recognition (CVPR)}, June
  2015.

\bibitem{johnson2018}
J.~Johnson, A.~Gupta, and L.~Fei-Fei, ``Image generation from scene graphs,''
  in {\em Proceedings of the IEEE Conference on Computer Vision and Pattern
  Recognition (CVPR)}, June 2018.

\bibitem{dhamo2020}
H.~Dhamo, A.~Farshad, I.~Laina, N.~Navab, G.~D. Hager, F.~Tombari, and
  C.~Rupprecht, ``Semantic image manipulation using scene graphs,'' in {\em
  CVPR}, 2020.

\bibitem{ji2020actiongenome}
J.~Ji, R.~Krishna, L.~Fei-Fei, and J.~C. Niebles, ``Action genome: Actions as
  compositions of spatio-temporal scene graphs,'' in {\em Proceedings of the
  IEEE/CVF Conference on Computer Vision and Pattern Recognition (CVPR)}, June
  2020.

\bibitem{wald20203dssg}
J.~Wald, H.~Dhamo, N.~Navab, and F.~Tombari, ``Learning 3d semantic scene
  graphs from 3d indoor reconstructions,'' in {\em Conference on Computer
  Vision and Pattern Recognition (CVPR)}, 2020.

\bibitem{armeni20193d}
I.~Armeni, Z.-Y. He, J.~Gwak, A.~R. Zamir, M.~Fischer, J.~Malik, and
  S.~Savarese, ``3d scene graph: A structure for unified semantics, 3d space,
  and camera,'' in {\em Proceedings of the IEEE International Conference on
  Computer Vision}, pp.~5664--5673, 2019.

\bibitem{rosinol2020dynamicsg}
A.~Rosinol, A.~Gupta, M.~Abate, J.~Shi, and L.~Carlone, ``{3D} dynamic scene
  graphs: Actionable spatial perception with places, objects, and humans,'' in
  {\em Robotics: Science and Systems (RSS)}, 2020.

\bibitem{mvor2020}
A.~Kadkhodamohammadi, A.~Gangi, M.~de~Mathelin, and N.~Padoy, ``A multi-view
  rgb-d approach for human pose estimation in operating rooms,'' in {\em 2017
  IEEE Winter Conference on Applications of Computer Vision (WACV)},
  pp.~363--372, 2017.

\bibitem{sharghi2020}
A.~Sharghi, H.~Haugerud, D.~Oh, and O.~Mohareri, ``Automatic operating room
  surgical activity recognition for robot-assisted surgery,'' in {\em Medical
  Image Computing and Computer Assisted Intervention -- MICCAI 2020} (A.~L.
  Martel, P.~Abolmaesumi, D.~Stoyanov, D.~Mateus, M.~A. Zuluaga, S.~K. Zhou,
  D.~Racoceanu, and L.~Joskowicz, eds.), (Cham), pp.~385--395, Springer
  International Publishing, 2020.

\bibitem{ransac1981}
M.~Fischler and R.~Bolles, ``Random sample consensus: A paradigm for model
  fitting with applications to image analysis and automated cartography,'' {\em
  Communications of the ACM}, vol.~24, no.~6, pp.~381--395, 1981.

\bibitem{icp1987}
K.~S. {Arun}, T.~S. {Huang}, and S.~D. {Blostein}, ``Least-squares fitting of
  two 3-d point sets,'' {\em IEEE Transactions on Pattern Analysis and Machine
  Intelligence}, vol.~PAMI-9, no.~5, pp.~698--700, 1987.

\bibitem{voxelpose}
H.~Tu, C.~Wang, and W.~Zeng, ``Voxelpose: Towards multi-camera 3d human pose
  estimation in wild environment,'' in {\em European Conference on Computer
  Vision (ECCV)}, 2020.

\bibitem{cheng2020bottom}
B.~Cheng, B.~Xiao, J.~Wang, H.~Shi, T.~S. Huang, and L.~Zhang, ``Higherhrnet:
  Scale-aware representation learning for bottom-up human pose estimation,'' in
  {\em CVPR}, 2020.

\bibitem{fasterrcnn2015}
S.~Ren, K.~He, R.~Girshick, and J.~Sun, ``Faster r-cnn: Towards real-time
  object detection with region proposal networks,'' in {\em Advances in Neural
  Information Processing Systems} (C.~Cortes, N.~Lawrence, D.~Lee, M.~Sugiyama,
  and R.~Garnett, eds.), vol.~28, Curran Associates, Inc., 2015.

\bibitem{dai2017scannet}
A.~Dai, A.~X. Chang, M.~Savva, M.~Halber, T.~Funkhouser, and M.~Nie{\ss}ner,
  ``Scannet: Richly-annotated 3d reconstructions of indoor scenes,'' in {\em
  Proc. Computer Vision and Pattern Recognition (CVPR), IEEE}, 2017.

\bibitem{fpfh2009}
R.~B. {Rusu}, N.~{Blodow}, and M.~{Beetz}, ``Fast point feature histograms
  (fpfh) for 3d registration,'' in {\em 2009 IEEE International Conference on
  Robotics and Automation}, pp.~3212--3217, 2009.

\bibitem{FORESTIER2012255}
G.~Forestier, F.~Lalys, L.~Riffaud, B.~Trelhu, and P.~Jannin, ``Classification
  of surgical processes using dynamic time warping,'' {\em Journal of
  Biomedical Informatics}, vol.~45, no.~2, pp.~255--264, 2012.

\bibitem{ahmadi2006}
S.-A. Ahmadi, T.~Sielhorst, R.~Stauder, M.~Horn, H.~Feussner, and N.~Navab,
  ``Recovery of surgical workflow without explicit models,'' in {\em Medical
  Image Computing and Computer-Assisted Intervention -- MICCAI 2006}
  (R.~Larsen, M.~Nielsen, and J.~Sporring, eds.), (Berlin, Heidelberg),
  pp.~420--428, Springer Berlin Heidelberg, 2006.

\bibitem{bellman1959}
R.~{Bellman} and R.~{Kalaba}, ``On adaptive control processes,'' {\em IRE
  Transactions on Automatic Control}, vol.~4, no.~2, pp.~1--9, 1959.

\bibitem{bagnall2017}
A.~Bagnall, J.~Lines, A.~Bostrom, J.~Large, and E.~Keogh, ``The great time
  series classification bake off: A review and experimental evaluation of
  recent algorithmic advances,'' {\em Data Min. Knowl. Discov.}, vol.~31,
  p.~606–660, May 2017.

\end{thebibliography}

\end{document}